\documentclass{article}
\usepackage{spconf,amsmath,graphicx}
\usepackage{amssymb,bm}
\usepackage{url}

\title{Overcomplete Representations Against Adversarial Videos}
\name{Shao-Yuan Lo, Jeya Maria Jose Valanarasu, and  Vishal M. Patel \thanks{This work was supported by the DARPA GARD Program HR001119S0026-GARD-FP-052.}}
\address{Dept. of Electrical and Computer Engineering, Johns Hopkins University, MD, USA \\\texttt{\{sylo, jvalana1, vpatel36\}@jhu.edu}}

\begin{document}
%
\maketitle
\begin{abstract}
Adversarial robustness of deep neural networks is an extensively studied problem in the literature and various methods have been proposed to defend against adversarial images.  However, only a handful of defense methods have been developed for defending against attacked videos. In this paper, we propose a novel Over-and-Under complete restoration network for Defending against adversarial videos (OUDefend). Most restoration networks adopt an encoder-decoder architecture that first shrinks spatial dimension then expands it back. This approach learns undercomplete representations, which have large receptive fields to collect global information but overlooks local details. On the other hand, overcomplete representations have opposite properties.  Hence, OUDefend is designed to balance local and global features by learning those two representations. We attach OUDefend to target video recognition models as a feature restoration block and train the entire network end-to-end. Experimental results show that the defenses focusing on images may be ineffective to videos, while OUDefend enhances robustness against different types of adversarial videos, ranging from additive attacks, multiplicative attacks to physically realizable attacks. Code: \url{https://github.com/shaoyuanlo/OUDefend}
\end{abstract}

\section{Introduction}

\begin{figure*}[htp!]
	\begin{center}
		\centering
		\includegraphics[width=0.76\textwidth]{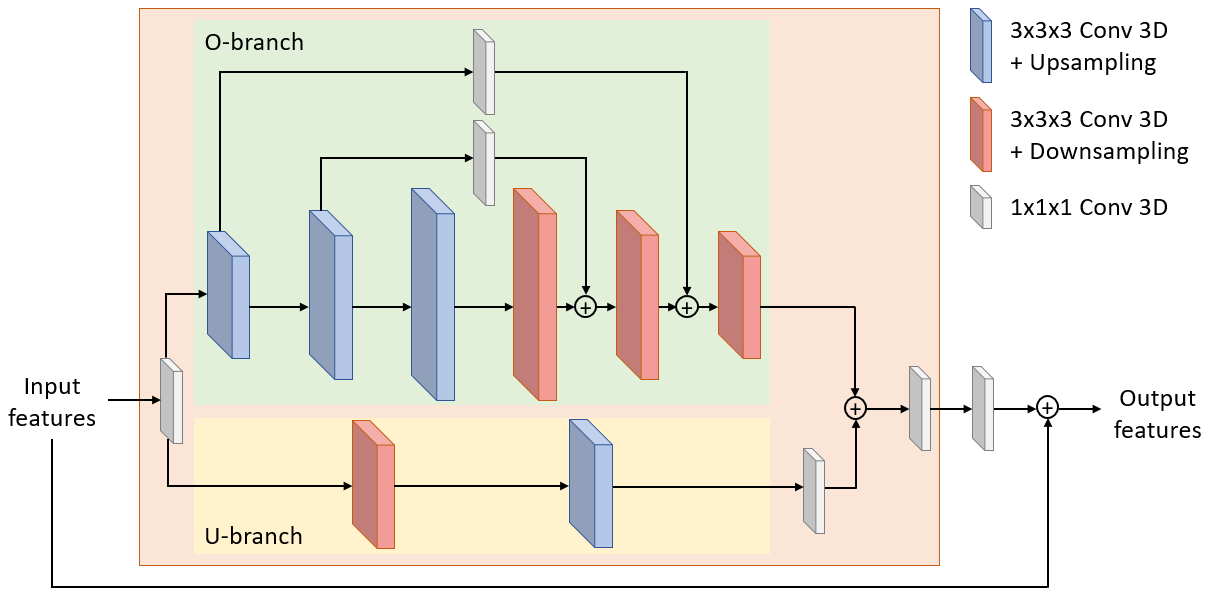}
		\caption{The proposed OUDefend architecture.}
		\label{fig:ounet}
	\end{center}
\end{figure*}

Despite achieving great success in many applications, deep neural networks (DNNs) have shown its vulnerability to adversarial examples \cite{Szegedy2014Intriguing,goodfellow2015explaining}. Thus, effective defense strategies are imperative. Several studies employ denoising-based methods to increase adversarial robustness. One stream uses denoising at the pre-processing stage to remove adversarial perturbations. HGD \cite{liao2018defense} adopts a U-Net-based \cite{ronneberger2015u} denoiser with high-level feature guidance against adversaries at the pixel level. ComDefend \cite{jia2019comdefend} mitigates adversarial effects through an image compression-based denoiser. However, this stream is easily defeated by the adaptive attacks \cite{obfuscated-gradients}. 

Other methods propose robust architectures, which contain denoising in network design instead of pre-processing and combine it with adversarial training. Adversarial training \cite{goodfellow2015explaining,madry2018towards} shows good adversarial robustness in the white-box setting and has been used as the foundation for defense. It requires a larger network capacity than standard training \cite{xie2020intriguing}, so designing network architectures having a high capacity to handle the difficult adversarial training is important. Xie et al. \cite{xie2019feature} developed a feature denoising block for their robust architecture design, which learns to remove perturbations at the feature level. Their method is only tested against a single attack type and only on the image data.

On the other hand, defenses in the video domain are less explored, where we are aware of only three studies for detection or defense against video attacks. AdvIT \cite{Xiao2019AdvITAF} detects adversarial video frames via temporal consistency, but provides no defense against the adversaries. Jia et al. \cite{jia2019identifying} introduced a similar detector, along with spatial and temporal defenses. However, these defenses are only tested in the black-box setting and thus their resistance to the stronger white-box attacks is not clear. Lo and Patel \cite{lo2020defending} leveraged multiple batch normalization branches for multi-perturbation robustness.

In this paper, we follow the robust architecture stream and propose a novel network, Over-and-Under complete restoration network for Defending against adversarial videos (OUDefend). OUDefend learns the \textit{overcomplete} representations \cite{lewicki2000learning} of input data against adversarial examples. Recall that Xie's method \cite{xie2019feature} considers several classical denoising operations, including non-local means \cite{buades2005non}, bilateral filters \cite{tomasi1998bilateral}, mean filters and median filters. Instead, we look into DNN-based algorithms for further improvements. 

Traditionally, image/video restoration networks adopt an encoder-decoder architecture where the encoder first downsamples the input to a lower dimension spatially and the decoder then upsamples it back to the original dimension \cite{yasarla2019uncertainty,zhang2018densely}. Here, the receptive field of the filters in the deeper layers gets enlarged. Such mechanism acquires \textit{undercomplete} representations, which focus more on high-level features and global information but pay less attention to local details. In contrast, overcomplete representations are good at extracting meaningful low-level features and local information that are favorable for restoration \cite{Jose_miccai2020}. Therefore, OUDefend consists of overcomplete and undercomplete braches to learn these two representation types respectively, and fuses their complementary features. We include OUDefend in target models as a feature restoration block and adversarially train the entire network end-to-end. The proposed method improves adversarial robustness against many different types of adversarial videos, including $\ell_\infty$-norm PGD \cite{madry2018towards}, $\ell_2$-norm PGD, multiplicative adversarial videos (MultAV) \cite{lo2020multav}, rectangular occlusion attack (ROA) \cite{Wu2020Defending}, adversarial framing (AF) \cite{zajac2019adversarial} and salt-and-pepper attack (SPA) \cite{lo2020defending}. In particular, we show that Xie's method \cite{xie2019feature} is ineffective when applied to video data.

\section{Proposed Method}

\setlength{\tabcolsep}{8pt}
\begin{table*}[!htbp]
	\begin{center}
		\caption{Evaluation results (\%) on UCF101. Rows are defense methods, and columns are the number of model parameters, clean input data and different attacks. Clean Model is trained on clean data, and the others are trained on clean data or a specific attack (corresponding to the columns). Avg$_{adv}$ is the average accuracy over the six attack types.}		
		\label{table:result}
		\begin{tabular}{l | r | r | rrrrrr | r}
			\hline \noalign{\smallskip} \noalign{\smallskip}
			Method & Params & Clean & PGD-$\ell_\infty$ & PGD-$\ell_2$ & MultAV & ROA & AF & SPA & Avg$_{adv}$ \\
			\noalign{\smallskip} \hline \noalign{\smallskip}
			Clean Model & 33.0M & 76.90 & 2.56 & 3.25 & 7.19 & 0.16 & 0.24 & 4.39 & 2.97 \\
			\noalign{\smallskip} \hline \noalign{\smallskip}
			Madry's method \cite{madry2018towards} & 33.0M & 76.90 & 33.94 & 35.05 & 47.00 & 41.29 & 74.81 & 55.99 & 48.01 \\
			Xie's method-A \cite{xie2019feature} & 33.7M & 70.82 & 31.48 & 33.25 & 42.69 & 37.59 & 58.87 & 49.14 & 42.17 \\
			Xie's method-B \cite{xie2019feature} & 34.8M & 69.47 & 30.19 & 32.65 & 41.87 & 38.22 & 58.74 & 49.14 & 41.80 \\			
			\noalign{\smallskip} \hline \noalign{\smallskip}
			OUDefend (ours) & 33.6M & \textbf{77.90} & \textbf{34.18} & \textbf{35.32} & \textbf{47.63} & \textbf{42.00} & \textbf{81.76} & \textbf{56.25} & \textbf{49.52} \\
			\noalign{\smallskip} \hline
		\end{tabular}
	\end{center}
\end{table*}

We use video restoration to develop a robust architecture that has innate adversarial robustness for the problem of defense against adversarial videos. Previous DNNs used for restoration adopt a generic encoder-decoder architecture in which the encoder extracts an abstract version of input data while removing noise \cite{yasarla2019uncertainty,zhang2018densely}. To elaborate, they employ convolutional layers followed by max-pooling layers in the encoder and upsampling layers in the decoder. Such architecture is an example of undercomplete DNNs because the spatial dimension of the latent space representation is smaller than the inputs. As the receptive fields of filters increase after every max-pooling layer, the learned undercomplete representations collect more high-level features and global context.

Overcomplete representations, in contrast to the undercomplete representations, were used as an alternative generic method for the representation of signals \cite{lewicki2000learning}. It involves using overcomplete dictionaries so that the number of basis functions is more than the number of input signal samples. This enables higher flexibility, leading to a robust representation of signals. Interestingly, DNNs employing overcomplete representations have not been explored much \cite{Jose_miccai2020}. The overcomplete representations of visual data are able to acquire more meaningful low-level features and local context which are favorable for video restoration. As a result, we design an overcomplete network architecture to exploit the overcomplete representations where we project the input to a higher dimension spatially. In our overcomplete network, the receptive field gets constrained and so more low-level features and fine details are learned even in the deep layers compared to an undercomplete network. This happens as we use upsampling layers after each convolutional layer (instead of max-pooling in undercomplete networks) in the encoder which prevents the receptive field from enlarging in the deep layers. Furthermore, we fuse the over-and-under complete representations to fully gain their complementary advantages.

The architecture of the proposed OUDefend is illustrated in Fig. \ref{fig:ounet}. It has two branches: an overcomplete branch (O-branch) and an undercomplete branch (U-branch). O-branch has six $3 \! \times \! 3 \! \times \! 3$ convolutional layers in total, where the encoder and decoder both have three layers each. In the encoder of O-branch, each convolutional layer is followed by an upsampling layer, whereas in the decoder, each convolutional layer is followed by a downsampling layer. We employ interpolation for upsampling and max-pooling for downsampling. Skip connections \cite{ronneberger2015u} are used between the encoder and decoder for forwarding the features from early layers to the later layers thus helping in efficient gradient propagation. In addition, we propose using a $1 \! \times \! 1 \! \times \! 1$ convolutional layer in each skip connection. This layer learns and decides the most efficient earlier layer features that should be fused with the features of later layers. Although O-branch learns overcomplete representations that capture better low-level features than undercomplete representations, we note that undercomplete representations are also necessary as they leverage some high-level feature information which improves feature denoising. Thus, we propose having U-branch, a standard encoder-decoder structure \cite{ronneberger2015u,badrinarayanan2017segnet} with downsampling in the encoder and upsampling in the decoder. As U-branch is used as an auxiliary branch in OUDefend, we make it a lightweight 2-layer structure to reduce computational cost.

Next, we integrate the features from O-branch and U-branch. Before fusion, we have a $1 \! \times \! 1 \! \times \! 1 $ convolutional layer at the end of U-branch to adjust the ratio of undercomplete representations before fusing them with their overcomplete counterparts.  We then increase the number of channels back by using an another $1 \! \times \! 1 \! \times \! 1 $ convolutional layer after the feature fusion of these two branches, and the outputs of this layer are the final restored feature maps. Finally, a $1 \! \times \! 1 \! \times \! 1$ convolutional layer and a residual connection are used. Since restoration may affect signals, it can maintain a balance between signal protection and noise suppression \cite{xie2019feature}. Furthermore, in order to keep the computational complexity low, at the beginning we pass the input features through a common $1 \! \times \! 1 \! \times \! 1$ convolutional layer to reduce the number of channels prior to feeding them to the two separate branches. All of the operations are 3D versions for processing video data.

\section{Experiments}

We evaluate the proposed OUDefend on six different types of adversarial videos: PGD-$\ell_\infty$ \cite{madry2018towards}, PGD-$\ell_2$, MultAV-$\ell_\infty$ \cite{lo2020multav}, ROA \cite{Wu2020Defending}, AF \cite{zajac2019adversarial} and SPA \cite{lo2020defending}. These attack approaches range from additive attacks, multiplicative attacks to physically realizable attacks, so we can thoroughly test the adversarial robustness of OUDefend. We also present some analysis of the method by displaying the feature maps under attacks.

\subsection{Experimental Setup}

\begin{figure*}[!htbp]
	\begin{center}
		\centering
		\includegraphics[width=0.89\textwidth]{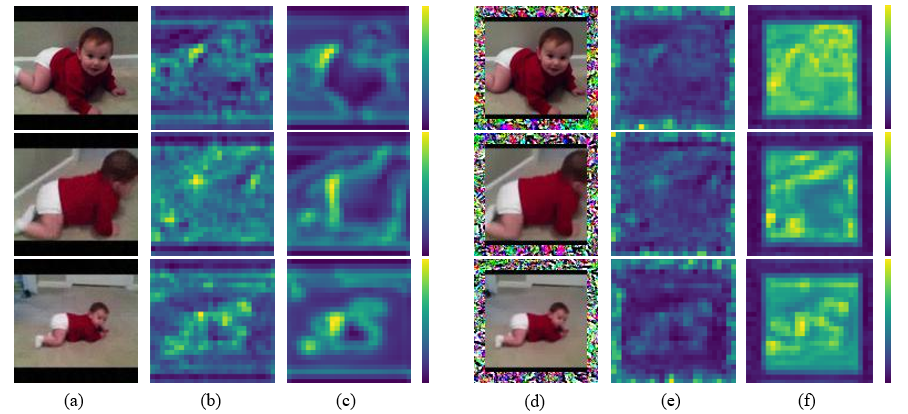}
		\caption{Feature maps after the conv2 block of Clean Model and OUDefend under PGD-$\ell_\infty$ and AF. Clean Model is vanilla 3D ResNet-18 trained on clean data. OUDefend is adversarially trained, and here it is inserted after the conv2 block. Top to bottom: Three selected frames from a video. (a) PGD-$\ell_\infty$ example. (b) Clean Model's features under PGD-$\ell_\infty$. (c) OUDefend's features under PGD-$\ell_\infty$. (d) AF example. (e) Clean Model's features under AF. (f) OUDefend's features under AF.}
		\label{fig:feature}
	\end{center}
\end{figure*}

Our experiments are performed on the UCF101 dataset \cite{Soomro2012ucf}, an action recognition dataset composed of 13,320 videos with 101 action classes. 3D ResNet-18 \cite{hara3dcnns}, a 3D convolution version of ResNet-18 \cite{he2016deep}, is adopted as our backbone network. We attach an OUDefend architecture to 3D ResNet-18 as a feature restoration block after its conv4 block. All the networks are trained end-to-end using SGD optimizer. For adversarial training, we follow Madry's protocol \cite{madry2018towards}.

For both inference and adversarial training, the settings of the six considered attacks are as follows \cite{lo2020defending}:
\begin{itemize}
	\item PGD-$\ell_\infty$: $\epsilon=4/255$, $\alpha=1/255$, and $T=5$.
	\item PGD-$\ell_2$: $\epsilon=160$, $\alpha=1.0$, and $T=5$.
    \item MultAV-$\ell_\infty$: $\epsilon_m=1.04$, $\alpha_m=1.01$, and $T=5$.
	\item ROA: Rectangle size 30$\times$30, $\epsilon=255/255$, $\alpha=70/255$, and $T=5$.
	\item AF: Framing width 10, $\epsilon=255/255$, $\alpha=70/255$, and $T=5$.
	\item SPA: 100 adversarial pixels on each video frame, $\epsilon=255/255$, $\alpha=70/255$, and $T=5$.
\end{itemize}

We compare our OUDefend with Madry's method \cite{madry2018towards} and Xie's method \cite{xie2019feature}. Here, Madry's method is equivalent to the vanilla 3D ResNet-18 with adversarial training. Xie's method adds four feature denoising blocks to ResNet after the conv2, conv3, conv4 and conv5 blocks. It performs the same adversarial training protocol as Madry's method. We compare with their Gaussian version of non-local means denoising, which has two $1 \! \times \! 1$ convolutional layers for embeddings. It is their best performing denoising operation. Since Xie's method is designed for only image data, we extend it to the video domain in two ways: Xie's method-A which replaces its operations by 3D versions directly, and Xie's method-B which conducts the original 2D operations on videos frame-by-frame.

\subsection{Evaluation Results}

Table \ref{table:result} reports our experimental results on the UCF101 video recognition dataset. Both Xie's method-A and Xie's method-B fail to improve Madry's method.  In fact, these methods' performance on the clean data drops as well. This indicates that their denoising structure does not work on video data and might degrade the quality of features. The reason may be that conventional image denoising operations cannot be generalized to video denoising very well, particularly when they are included as a part of a deep learning model, i.e., they are not compatible with video DNNs. Instead, OUDefend applies to 3D convolutional network architectures. It achieves the best adversarial robustness across all the six attack approaches we consider, showing its effectiveness ranges from additive attacks, multiplicative attacks to physically realizable attacks. Moreover, OUDefend's clean data performance is also better than the baseline architecture and Xie's method, which demonstrates that adding OUDefend as a restoration block will not degrade the feature quality.

To show the importance of learning overcomplete representations in OUDefend, we build a variant which has only U-branch and thus learns undercomplete representations only. Under PGD-$\ell_\infty$ attack, this variant obtains 33.15\% accuracy, lower than OUDefend. This demonstrates the advantage of learning overcomplete representations.

On the other hand, OUDefend is a lightweight architecture that has only 0.6M parameters. It just accounts for 1.8\% number of parameters when it is deployed in 3D ResNet-18. It achieves the best performance and robustness with a fewer number of parameters than both Xie's method-A and Xie's method-B.

\subsection{Feature Map Visualization}

We visualize the feature maps of Clean Model and OUDefend under the PGD-$\ell_\infty$ and AF attacks in Fig. \ref{fig:feature}. As it can be seen, Clean Model's features under PGD-$\ell_\infty$ are noisy. Their activations are scattered over semantically trivial areas and thus fail to focus on informative content. The proposed OUDefend with adversarial training leads to clearer features capturing fine details. Specifically, adversarial training forces models to learn meaningful patterns in adversarial examples, and OUDefend further restores the perturbed features by leveraging over-and-under complete representations.

In the case of AF attacks, Clean Model is misled to focus on the border of the video frames where the adversarial framings are located. Apparantly, this area is semantically trivial. OUDefend with adversarial training learns to ignore the border area and pay attention to the semantically meaningful regions. We can observe that the activations at the border area are almost zero. Therefore, the effectiveness of the proposed method is demonstrated by feature visualization as well.


\section{Conclusion}

We propose OUDefend, a new robust network architecture that exploits overcomplete representations to restore adversarial features. With an auxiliary undercomplete representation branch, it is able to balance local and global context by fusing these two representations. Adversarial robustness in the video domain is less explored, and our experiments show that the defenses tailored to images may be ineffective to videos. In contrast, OUDefend enhances robustness to many different types of adversarial videos, ranging from additive attacks, multiplicative attacks to physically realizable attacks.

\bibliographystyle{IEEEbib}
\bibliography{cite_oudefend}

\end{document}